\documentclass{bmvc2k}

\title{ACIL: Active Class Incremental Learning for Image Classification}

\addauthor{Aditya R. Bhattacharya$^*$}{arb17b@fsu.edu}{1}
\addauthor{Debanjan Goswami$^*$}{dgoswami@fsu.edu}{1}
\addauthor{Shayok Chakraborty}{shayok@cs.fsu.edu}{1}

\addinstitution{
 Department of Computer Science\\
 Florida State University\\
 Tallahassee, Florida, USA
}
\runninghead{ACIL}{Active Class Incremental Learning for Image Classification}

\usepackage{graphicx}
\usepackage{amsmath}
\usepackage{amssymb}
\usepackage{booktabs}
\usepackage{subfigure}
\usepackage{graphicx}
\usepackage{amsmath}
\usepackage{amsthm}
\usepackage{booktabs}
\usepackage{algorithm}
\usepackage{algorithmic}
\usepackage{wrapfig}
\usepackage[switch]{lineno}

\DeclareMathOperator*{\argmin}{arg\,min}

\def\etal{\emph{et al}\bmvaOneDot}

\begin{document}

\maketitle

\def\thefootnote{*}\footnotetext{These authors contributed equally to this work}

\begin{abstract}
Continual learning (or class incremental learning) is a realistic learning scenario for computer vision systems, where deep neural networks are trained on episodic data, and the data from previous episodes are generally inaccessible to the model. Existing research in this domain has primarily focused on avoiding catastrophic forgetting, which occurs due to the continuously changing class distributions in each episode and the inaccessibility of the data from previous episodes. However, these methods assume that all the training samples in every episode are annotated; this not only incurs a huge annotation cost, but also results in a wastage of annotation effort, since most of the samples in a given episode will not be accessible to the model in subsequent episodes. Active learning algorithms identify the salient and informative samples from large amounts of unlabeled data and are instrumental in reducing the human annotation effort in inducing a deep neural network. In this paper, we propose ACIL, a novel active learning framework for class incremental learning settings. We exploit a criterion based on uncertainty and diversity to identify the exemplar samples that need to be annotated in each episode, and will be appended to the data in the next episode. Such a framework can drastically reduce annotation cost and can also avoid catastrophic forgetting. Our extensive empirical analyses on several vision datasets corroborate the promise and potential of our framework against relevant baselines. 
\end{abstract}

%-------------------------------------------------------------------------
\section{Introduction}
\label{sec_intro}

Deep neural networks (DNNs) have pushed the boundaries of computer vision and have achieved state-of-the-art performance in a variety of applications \cite{1_survey, 2_survey, 3_survey}. %Typically, a large data corpus is collected first (such as images \cite{4_survey} or text \cite{5_survey}) and the deep network is then trained on the pre-collected dataset over multiple epochs. However, 
In many real-world applications, the training data arrives in episodes over time, with each episode containing samples from a different set of classes. Further, this episodic data is ephemeral and cannot be held for long due to storage and privacy constraints \cite{8_survey, 9_survey, 11_survey}. The DNN is expected to learn incrementally about the new classes, from their episodic data; at the same time, it should retain knowledge about the formerly learned classes in previous episodes, whose data is no longer accessible. This learning paradigm is called \textit{Class Incremental Learning (CIL)}, which aims to continually induce a holistic classifier among all the classes encountered. A problem is CIL is \textit{catastrophic forgetting}, which occurs when optimizing the network to recognize the new classes erases its knowledge about the former classes (in previous episodes), resulting in a degradation in performance. Thus, developing algorithms to mitigate catastrophic forgetting has been the primary focus in CIL research \cite{CIL_survey}. 

However, another aspect that has been overlooked in this context is the cost of data annotation; CIL algorithms assume that all the training samples in each episode are annotated with their class labels. This incurs a huge cost as data annotation is an expensive process in terms of time, labor and human expertise. Moreover, this results in a wastage of annotation effort, since most of the data samples in a given episode will not be accessible to the deep model in subsequent episodes. This necessitates a strategy to more efficiently utilize the human annotation effort in CIL settings. 

\textit{Active Learning (AL)} algorithms have gained popularity in reducing the human annotation effort in inducing a machine learning / deep learning model \cite{Settles_2010}. When faced with large amounts of unlabeled data, these algorithms automatically identify the salient and the most informative samples that need to be annotated manually. AL has been successfully used in a variety of applications, such as computer vision \cite{Yoo_2019}, text mining \cite{tong_support_2000}, medical diagnosis \cite{Wu_2021} and bioinformatics \cite{Hatice_2010} among others. 

In this paper, we introduce \textit{ACIL}, a novel \textbf{A}ctive learning framework for \textbf{C}lass \textbf{I}ncremental \textbf{L}earning. Contrary to existing CIL settings, we assume that in each episode, only a small portion of the data is annotated; majority of the samples are unlabeled in each episode. In each episode, we are allowed to select an exemplar set of size $k$ (where $k$ is a pre-determined number) which will be manually annotated and appended to the data in the next episode. The exemplar set can contain samples from the labeled and unlabeled sets in the current episode as well as samples from the exemplar set obtained from the previous episode. We formulate a criterion based on diversity and uncertainty and use weighted set partitioning techniques to select the set of exemplar samples. %A deep neural network (DNN) is trained from scratch in each episode. 
Such a framework can drastically reduce the data annotation cost, as only a small fraction of the samples is annotated in each episode. It can also potentially avoid catastrophic forgetting, as the exemplar samples in each episode are propagated to subsequent episodes. %To the best of our knowledge, such a learning setup has not been explored in the CIL literature. 

%We present a survey of related research in Section \ref{sec_related}; the proposed ACIL framework is detailed in Section \ref{sec_proposed}; the results of our empirical studies are presented in Section \ref{sec_expt}; and we conclude with discussions in Section \ref{sec_conc}. 

%----------------------------------------------------------------------------------------------------------------

\section{Related Work}
\label{sec_related}

%We present a survey of existing research in class incremental learning and active learning in this section. 

\hspace{.2in} \textbf{Class Incremental Learning:} Class Incremental Learning (CIL) has attracted significant research attention in recent years in the computer vision and machine learning communities; please refer to \cite{CIL_survey} for a detailed taxonomy and survey. CIL can be broadly organized into three categories: \textit{data-centric, model-centric} and \textit{algorithm-centric}. Data-centric CIL methods attempt to alleviate catastrophic forgetting with the help of extra data, called exemplars, and are further categorized into data replay and data regularization. Replay based methods attempt to overcome catastrophic forgetting by revisiting exemplar samples from past episodes; these methods sample an exemplar set from each episode, which is appended to the data in the following episode. In each episode, the deep model is trained using the data from that episode, as well as the exemplars obtained from previous episodes \cite{39_survey, 40_survey}. Several strategies have been explored to select the exemplar set, such as \textit{Rainbow Memory} \cite{35_survey} which selects uncertain samples given by the variance of the model predictions on a pertubed set of samples; \textit{GDumb} \cite{GDumb_Paper} which greedily selects samples with the constraint to asymptotically balance the class distribution of the selected samples; and \textit{iCaRL} \cite{32_survey} which iteratively constructs the exemplar set such that the newly added exemplar causes the average feature vector over all the exemplars to best approximate the average feature vector over all the training examples. Generative replay based methods have also been studied, which attempt to model the distribution of data in former episodes and generate samples accordingly \cite{48_survey, 49_survey}. %Another strategy of data-centric CIL is to regularize the model with past data and control the optimization direction \cite{62_survey, 63_survey}.   
Model-centric methods either regularize the model parameters or expand the network structure for stronger representation ability \cite{140_survey, 141_survey}. Algorithm-centric CIL methods focus on designing algorithms to maintain the model's knowledge in former tasks to mitigate catastrophic forgetting. Knowledge distillation-based CIL methods fall in this category, where the goal is to build a mapping between old and new models and transfer the knowledge of the old model in inducing the new model \cite{85_survey, 159_survey, 160_survey}.

\textbf{Active Learning:} Active Learning (AL) is a well-studied problem in the machine vision literature \cite{Settles_2010}. %The most common query strategy in AL is uncertainty sampling, where unlabeled samples with the highest prediction uncertainties are queried for their labels \cite{Joshi_2012, Guo_2013}. 
With the popularity of deep neural networks, \textit{deep active learning (DAL)} has attracted research attention, where the goal is to query informative unlabeled samples for manual annotation and simultaneously learn discriminating feature representations using a deep neural network \cite{Ren_2021}. Common DAL techniques include a task agnostic scheme which learns a loss prediction function to predict the loss value of an unlabeled sample and queries samples accordingly \cite{Yoo_2019}, a greedy technique to query a coreset of samples that represent the whole dataset \cite{Coreset_Paper}, a sampling technique based on diverse gradient embeddings \textit{(BADGE)} \cite{Badge_Paper}, a strategy based on temporal output discrepancy that queries samples based on the discrepancy of outputs given by the models at different optimization steps during training \cite{TOD_ICCV_2021} and an AL framework that queries unlabeled samples that can provide the most positive influence on model performance \cite{liu2021influence}. Techniques based on adversarial learning have depicted particularly impressive performance in DAL \cite{Sinha_2019, Zhu_2017, Ducoffe_2018}. 

Even though both CIL and AL have been well-researched, their combination has received significantly less attention. A body of research has focused on updating the deep model on-the-fly (rather than retraining from scratch) with actively sampled training data \cite{brust2020active}; this setup assumes that the entire training and unlabeled sets are accessible to the deep model throughout the process, which is different from the classical CIL setting. Active class selection has also been studied in the context of CIL, where the incremental learner selects the classes to receive additional training instances from \cite{mcclurg2023active}; this is also different from the classical CIL setting, where the learner has no control over the classes of the samples arriving in each episode. Belouadah \etal studied the performance of AL algorithms in a CIL setup \cite{belouadah2020active}. %However, this method assumes all the samples are annotated in the first episode, which may restrict its application; \textit{ACIL} does not make this assumption and is thus more realistic. We compare the performance of \textit{ACIL} against this method with two state-of-the-art AL techniques, \textit{Coreset} and \textit{BADGE}, in Section \ref{sec_expt}. 
%The learning setup in \cite{belouadah2020active} is similar to that of \textit{ACIL}. 
However, \textit{ACIL} differs from \cite{belouadah2020active} in the following two fundamental ways: $(i)$ \cite{belouadah2020active} assumes all the samples are annotated in the first episode, which may restrict its application; \textit{ACIL} does not make this assumption and is thus more realistic. $(ii)$ The active sampling criterion in \cite{belouadah2020active} is applied only on the unlabeled data in each episode to create the exemplar set; in contrast, \textit{ACIL} samples intelligently from the unlabeled set and also the exemplar set obtained from the previous episode, and thus addresses catastrophic forgetting more efficiently. This has been empirically validated in Section \ref{sec_expt}, where \textit{ACIL} has been compared against this method with two state-of-the-art AL techniques: \textit{Coreset} and \textit{BADGE}. 

%----------------------------------------------------------------------------------------------------------------

\section{Proposed Framework}
\label{sec_proposed}

\subsection{Problem Setup}

\begin{figure*}[h]
  \centering
 % \fbox{\rule{0pt}{2in} \rule{0.9\linewidth}{0pt}}
   \includegraphics[width=0.85\linewidth]{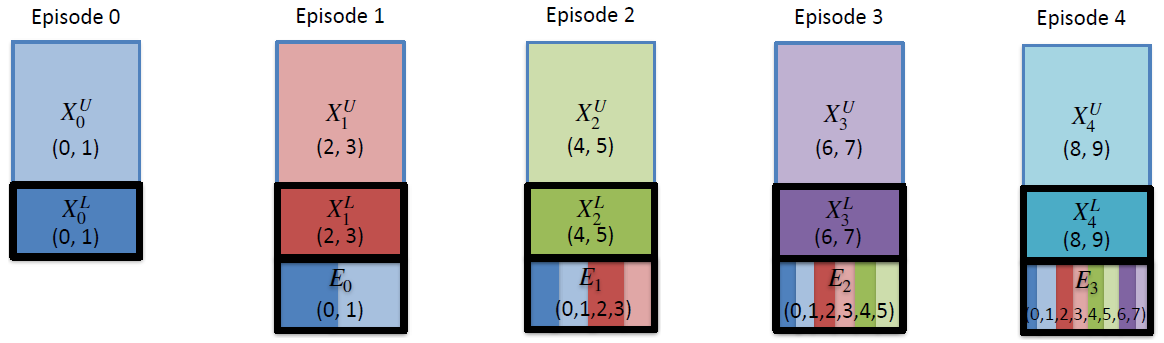}
   \caption{Learning setup of \textit{ACIL}. The numbers in parentheses denote the class labels present in the corresponding set. We consider the disjoint CIL setup \cite{35_survey}, where the samples is a given episode belong to different classes compared to the samples in all the previous episodes. Please refer to the text for more details.}
   \label{fig_setup}
\end{figure*}

The setup of \textit{ACIL} is depicted in Figure \ref{fig_setup}. The training data arrives sequentially over $N$ episodes (numbered $0, 1, \ldots N-1$); we consider the \textit{disjoint CIL} setup \cite{35_survey}, where the samples in a given episode belong to different classes compared to the samples in all the previous episodes. The data in a particular episode $n$ consists of two parts: a labeled part $X^{L}_{n}$ and an unlabeled part $X^{U}_{n}$ ($|X^{L}_{n}| \ll |X^{U}_{n}|$). Except episode $0$, each episode also contains a labeled exemplar set $E_{n-1}$, which contains a subset of samples from all the previous episodes. From each episode $n$, we are allowed to select an exemplar set $E_{n}$ of a fixed size $k$, containing a subset of samples from $X^{L}_{n}$, $X^{U}_{n}$ and $E_{n-1}$. The samples in the selected exemplar set $E_{n}$ are annotated and appended to the data in the next episode $(n+1)$. Note that, since $X^{L}_{n}$ and $E_{n-1}$ are already annotated, annotating the samples in the selected exemplar set $E_{n}$ amounts to annotating only the samples selected from $X^{U}_{n}$. In each episode $e$, a deep neural network is trained on the labeled data $X^{L}_{n}$ and $E_{n-1}$. Note that the size (budget) of the exemplar set remains constant throughout the learning process and does not increase with increasing number of episodes; this appropriately captures a real-world learning setup where the memory budget is fixed. \textit{Also note that the learning setup in ACIL is similar in spirit to replay based methods in data-centric CIL; however, these methods require all the samples to be annotated in each episode, which entails significant manual labor. In contrast, ACIL identifies an exemplar set in each episode without requiring all the samples to be annotated, and thus results in substantial savings in annotation effort}. 

In the following sections, we detail how we split the budget $k$ to select samples from the sets $X^{L}_{n}$, $X^{U}_{n}$ and $E_{n-1}$ in a given episode $n$, our strategy to actively select the samples from each set and the loss function used to train the deep neural network in each episode. 

\subsection{Budget Splitting Strategy}
\label{sec_budget_split}

Conventional AL algorithms select all the samples only from the unlabeled set. However, in our setup, selecting all the $k$ samples from the unlabeled set will only capture information about the data in the current episode; this will fail to capture any knowledge about the data in former episodes, which may result in catastrophic forgetting. We therefore also need to select samples from the exemplar set $E_{n-1}$, as it contains useful information about the data encountered in the previous episodes. Hence, we propose a split of the allowed budget $k$ to select samples both from the labeled exemplar set and the unlabeled set in each episode. 

Let $C_{exemplar}$ denote the set of classes in the exemplar set $E_{n-1}$; this can be easily obtained since the exemplar set is fully labeled. Also, let $C_{episode}$ denote the set of classes that are contained in the current episodic data; this can be obtained since a part of the episodic data $X^{L}_{n}$ is labeled (we assume $X^{L}_{n}$ contains at least one sample from each class that is present in the current episodic data). %Note that $C_{exemplar}$ and $C_{episode}$ are disjoint sets, as the current episode contains samples from classes that were not seen in any of the previous episodes (similar to a typical disjoint CIL setup). 
Our rationale is to split the budget $k$ between $X^{U}_{n}$ and $E_{n-1}$ in the same proportion as the number of classes in the two sets. Thus, the number of samples to be selected from the episodic unlabeled set $X^{U}_{n}$ is:
%\vspace{-3pt}
\begin{equation}
\label{eqn_budget_split_1}
k_{unlabeled} = \left( \frac{|C_{episode}|}{|C_{episode}| + |C_{exemplar}|} \right).k
%\vspace{-3pt}
\end{equation}

\noindent Similarly, the number of samples to be selected from the episodic exemplar set $E_{n-1}$ is:
%\vspace{-3pt}
\begin{equation}
\label{eqn_budget_split_2}
k_{exemplar} = \left( \frac{|C_{exemplar}|}{|C_{episode}| + |C_{exemplar}|} \right).k
%\vspace{-3pt}
\end{equation}

Such a split ensures that we have more budget allocated for the set that contains more classes. Further, our active sampling strategy (detailed below) is applied class-wise on $X^{U}_{n}$ and $E_{n-1}$. Thus, the budget for each class in $X^{U}_{n}$ is  $\frac{k_{unlabeled}}{|C_{episode}|}$ and that for each class in $E_{n-1}$ is $\frac{k_{exemplar}}{|C_{exemplar}|}$. Since the samples in $X^{U}_{n}$ are unlabeled, we use the pseudo-labels furnished by the deep neural network obtained in episode $n-1$ to select samples from each class in $X^{U}_{n}$. Such a budget allocation strategy ensures that the exemplar set $E_{n}$ selected from episode $n$ contains a good representation of all the classes seen so far; thus, when it is appended to the data in episode $(n+1)$, the deep model trained in episode $(n+1)$ will potentially mitigate catastrophic forgetting. We do not select any samples from $X^{L}_{n}$ in $E_{n}$, as $X^{L}_{n}$ contains the same classes as $X^{U}_{n}$ and since $|X^{U}_{n}| \gg |X^{L}_{n}|$, it most likely contains more informative samples than $X^{L}_{n}$. Thus, the budget for the current episode is completely allocated to $X^{U}_{n}$. 

\subsection{Active Sampling Strategy}
\label{sec_AL}

We use the same active sampling strategy to select samples from each class separately in the sets $X^{U}_{n}$ and $E_{n-1}$. In this section, we present the general algorithm to actively sample a subset of a given size from a given set of samples. Let $X = \{x_{1}, x_{2}, \ldots x_{L}\}$ denote a set of $L$ samples, from which we are tasked to select a batch of $B$ samples in our exemplar set. Let $\mathcal{F}(x_{i})$ denote the feature embedding of sample $x_{i}$, obtained from the trained DNN. We exploit a strategy based on uncertainty and diversity to select the $B$ samples. Specifically, we identify the diverse samples by partitioning the set $X$ into $B$ diverse sets using a partition function $\mathcal{P}: X \rightarrow \{X_{1}, X_{2}, \ldots X_{B}\}$. Let $\{c_{1}, c_{2}, \ldots c_{B}\}$ denote the corresponding centroid of each set. Our objective is to group similar instances in the feature space into a set $X_{b}$, $b = 1, 2, \ldots B$, that is, each partition $X_{b}$ should have low variance $\sigma^{2}(X_{b})$. Using the difference method \cite{48_CLUE}, the variance $\sigma^{2}(X_{b})$ of partition $X_{b}$ can be computed as:  
%\vspace{-6pt}
\begin{equation}
\label{eqn_variance}
\sigma^{2}(X_{b}) = \frac{1}{2|X_{b}|^{2}} \sum_{x_{i}, x_{j} \in X_{b}} ||\mathcal{F}(x_{i}) - \mathcal{F}(x_{j})||^{2}
%\vspace{-8pt}
\end{equation}

To incorporate the prediction uncertainty in sample selection, we reformulate the set partitioning problem to minimize the \textit{weighted variance} within each partition (instead of the simple variance), where each sample is weighted based on its prediction uncertainty \cite{28_CLUE}. The objective function for set partitioning using weighted variance can be posed as: 
%\vspace{-8pt}
\begin{equation}
\label{eqn_wtd_variance}
\argmin_{\mathcal{P}} \sum_{b=1}^{B}  \sum_{x \in X_{b}} \mathcal{I}(x) ||\mathcal{F}(x) - c_{b}||^{2}
%\vspace{-6pt}
\end{equation}

\noindent where $\mathcal{I}(x)$ denotes the prediction uncertainty of sample $x$, computed using the Shannon's entropy as:
%\vspace{-8pt}
\begin{equation}
\label{eqn_entropy}
\mathcal{I}(x) = - \sum_{i=1}^{C} p(x)_{i} \log p(x)_{i}
%\vspace{-8pt}
\end{equation}

\noindent where $p(x)_{i}$ denotes the probability of sample $x$ with respect to class $i$ and $C$ is the total number of classes. Entropy is a widely used measure of uncertainty in AL research \cite{Ren_2021}. 

The optimization problem in Equation (\ref{eqn_wtd_variance}) in NP-hard. We use the weighted $k$-means algorithm \cite{Wtd_k_means} to approximate the solution to this problem, where $k$ is taken to be equal to the budget $B$. After clustering, the sample closest to the weighted mean in each partition $X_{b}$ is selected in the exemplar set. Weighted $k$-means has been used to solve set partitioning problems with promising results \cite{Wtd_k_means_SP, Clue_Paper}. 

\subsection{Loss Function}

In each episode $n$, a deep neural network is trained on the labeled episodic data $X^{L}_{n}$ and the labeled exemplar set $E_{n-1}$ obtained from the previous episode. Since typically, $|X^{L}_{n}| \gg  |E_{n-1}|$, there may exist a class imbalance in the labeled set $X^{L}_{n} \cup E_{n-1}$. We therefore use a weighted cross entropy loss \cite{Ho_2019} on the labeled set; for a sample $x_{i}$ in the labeled set, the loss is computed as:
%\vspace{-8pt}
\begin{equation}
\label{eqn_wtd_CE}
\mathcal{L_{WCE}}(x_{i}) = - \sum_{j=1}^{C} \delta(y_{i} == j) w_{j} \log p_{ij}
%\vspace{-8pt}
\end{equation}

\noindent Here $\delta(.)$ is the indicator function (whose value is $1$ if the argument is true and $0$ otherwise), $p_{ij}$ denotes the probability of sample $x_{i}$ with respect to class $j$ obtained using the soft-max activation of the DNN. $w_{j}$ denotes the importance weight assigned to class $j$ and is inversely proportional to the number of samples in class $j$ in the labeled set: $w_{j} = \frac{\alpha}{n_{j}}$, where $n_{j}$ is the number of samples in class $j$. The total loss over all the labeled samples is given by: 
%\vspace{-8pt}
\begin{equation}
\label{eqn_wtd_CE_total}
\mathcal{L_{WCE}} = \frac{1}{|L|} \sum_{i=1}^{|L|} \mathcal{L_{WCE}}(x_{i}) 
%\vspace{-8pt}
\end{equation}

Further, knowledge distillation has proved to be effective in CIL, where knowledge from the deep model $\mathcal{M}_{n-1}$, trained in episode $n-1$, is distilled to train the deep model $\mathcal{M}_{n}$ in episode $n$. $\mathcal{M}_{n-1}$ is applied on the exemplar set $E_{n-1}$ and a distillation loss $\mathcal{L_{D}}$ is computed between the predictions furnished by $\mathcal{M}_{n}$ and $\mathcal{M}_{n-1}$ on $E_{n-1}$. Please refer to \cite{ICARL_Paper, KD_Ref} for details about computing the distillation loss in a CIL setting. The total loss to train the DNN in episode $n$ is thus:
%\vspace{-8pt}
\begin{equation}
\label{eqn_total}
\mathcal{L} =  \mathcal{L_{WCE}} + \lambda \mathcal{L_{D}}
%\vspace{-8pt}
\end{equation}

\noindent where $\lambda$ is a weight parameter controlling the relative importance of the two terms. The pseudo-code of \textit{ACIL} is provided in the Supplemental File. As evident from the pseudo-code, our algorithm is computationally lightweight, simple and easy to implement. 

%----------------------------------------------------------------------------------------------------------------

\section{Experiments and Results}
\label{sec_expt}

\hspace{.2in} \textbf{Datasets:} We used six computer vision datasets to study the performance of \textit{ACIL}: MNIST \cite{MNIST_dataset}, SVHN \cite{SVHN_dataset}, CIFAR 10 \cite{CIFAR_dataset}, CIFAR 100 \cite{CIFAR_dataset}, COIL \cite{COIL_dataset} and Tiny ImageNet \cite{TinyImageNet_dataset}. The sizes of the labeled set $X^{L}_n$, unlabeled set $X^{U}_{n}$ and exemplar set $E_{n}$ for each dataset are detailed in the Supplemental file. 

\textbf{Comparison Baselines:} We used methods from both IL and AL as comparison baselines in our work: $(i)$ \textbf{AL baselines:} We used $3$ active learning techniques as comparison baselines: \textit{Random Sampling}, which queries a batch of samples at random from the current episode; \textit{Coreset} \cite{Coreset_Paper}; and \textit{BADGE} \cite{Badge_Paper}. Both \textit{Coreset} and \textit{BADGE} are widely used AL algorithms \cite{Ren_2021}. Similar to \textit{ACIL}, the AL baselines do not require all the samples to be annotated in each episode and operate on $X^{L}_n$ and $X^{U}_{n}$. As proposed in \cite{belouadah2020active}, the samples in the exemplar set are selected from the unlabeled data only in each episode, by applying different AL techniques. $(ii)$ \textbf{CIL baselines:} We used the following CIL methods as baselines in our research: \textit{iCaRL} \cite{32_survey}; \textit{GDumb} \cite{GDumb_Paper}; \textit{Rainbow Memory} \cite{35_survey}; and \textit{Finetuning} \cite{CIL_survey}. Note that, \textit{iCaRL}, \textit{GDumb} and \textit{Rainbow Memory} are all recently proposed replay based CIL techniques which use an exemplar set in each episode to mitigate catastrophic forgetting; these were selected since \textit{ACIL} closely resembles a replay based CIL setting. \textit{Finetuning} is a control CIL baseline, which does not use an exemplar set, instead it uses the same model that was trained on the previous episode and incrementally updates it using the data in the current episode.  

\textbf{Evaluation Metrics:} We used the following metrics to evaluate the performance of our framework: $(i)$ \textit{Accuracy}, which depicts how well the deep model generalizes on test data containing classes seen in the current and all the previous episodes; after each episode $n$, this metric was computed as the average accuracy of the model on the test set of each episode $1, 2, \ldots, n$. $(ii)$ \textit{Annotation effort}, computed as the number of samples that had to be annotated in each episode. $(iii)$ \textit{Retention} (a metric for forgetting), which quantifies how much information the model retains from the first episode, and was computed as the accuracy of the model on the test set from the first episode. Accuracy and retention are commonly used metrics in CIL \cite{CIL_survey} while human annotation effort is used to quantify the performance of AL algorithms \cite{Ren_2021}. 

\textbf{Implementation Details:} We used ResNet-34 \cite{ResNet_Paper} as the backbone architecture in our work. The model was trained for $240$ epochs in each episode, with a batch size of $128$ and a learning rate of $0.001$, using the \textit{Adam} optimizer.

\subsection{Main Results}

\begin{figure*}[h]
	\centering
		\subfigure[MNIST]{
          \label{fig_mnist}
          \includegraphics[trim = 1.3in 3.2in 1.7in 3.4in,clip,width=.31\textwidth]{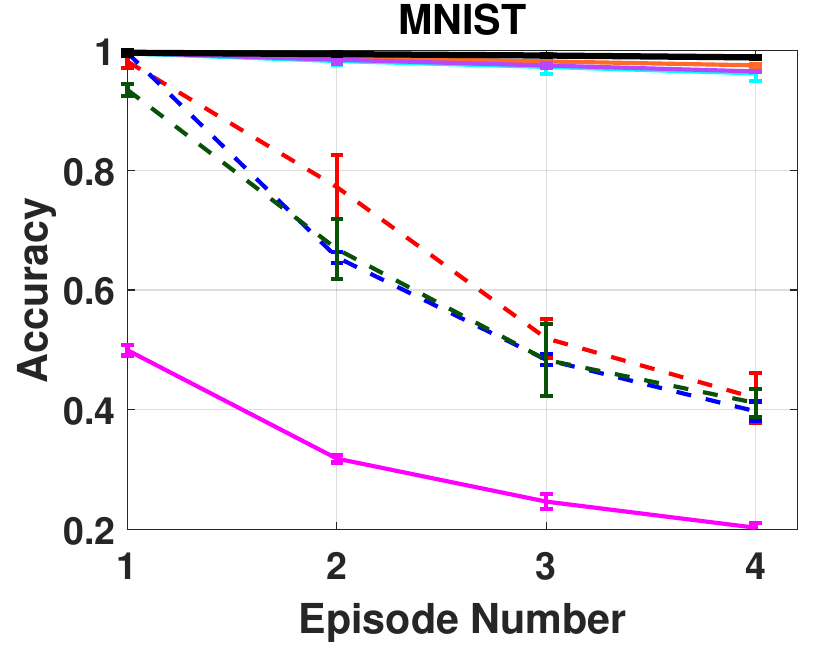}}   
     \hspace{.02in}
     \subfigure[CIFAR 10]{
          \label{fig_cifar10}
          \includegraphics[trim = 1.3in 3.2in 1.7in 3.4in,clip,width=.31\textwidth]{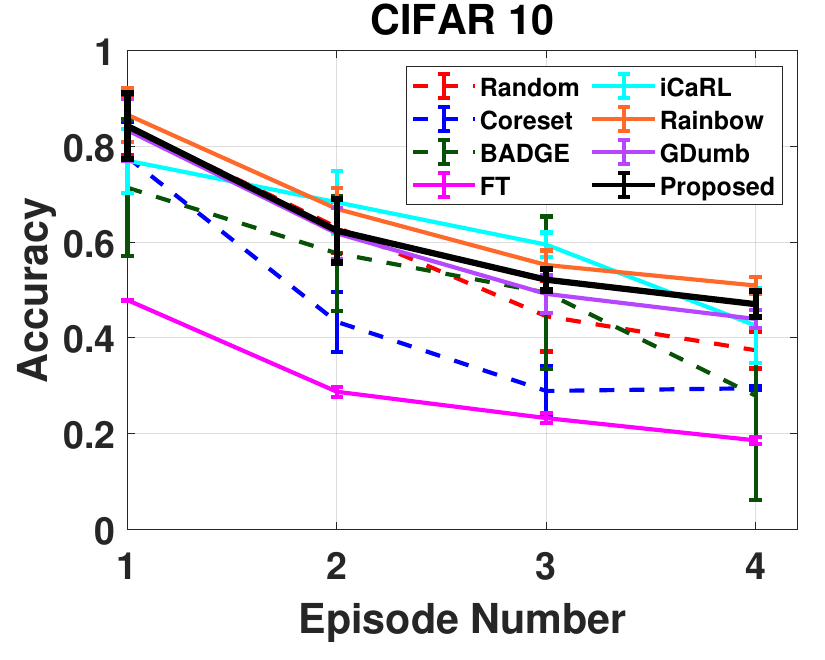}}
     \hspace{.02in}
		 \subfigure[SVHN]{
          \label{fig_svhn}
          \includegraphics[trim = 1.3in 3.2in 1.7in 3.4in,clip,width=.31\textwidth]{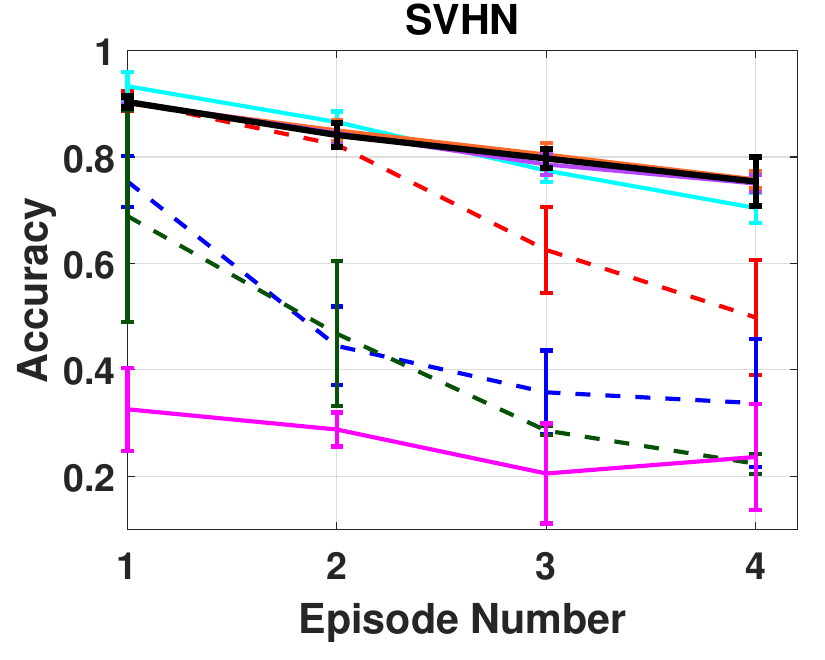}}    
          
           \vspace{.02in}
           
          \subfigure[COIL]{
          \label{fig_coil}
          \includegraphics[trim = 1.3in 3.2in 1.7in 3.4in,clip,width=.31\textwidth]{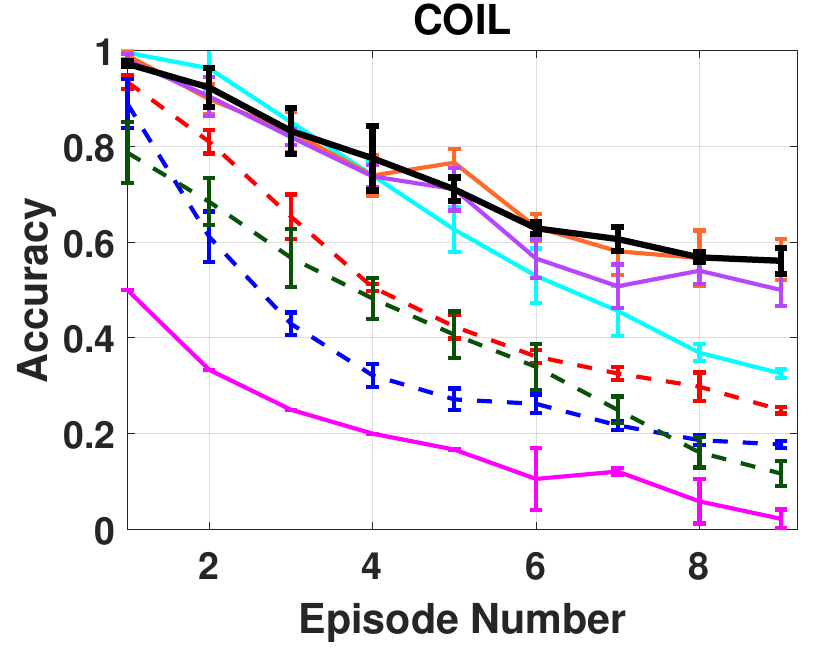}}   
     \hspace{.02in}
     \subfigure[CIFAR 100]{
          \label{fig_cifar100}
          \includegraphics[trim = 1.3in 3.2in 1.7in 3.4in,clip,width=.31\textwidth]{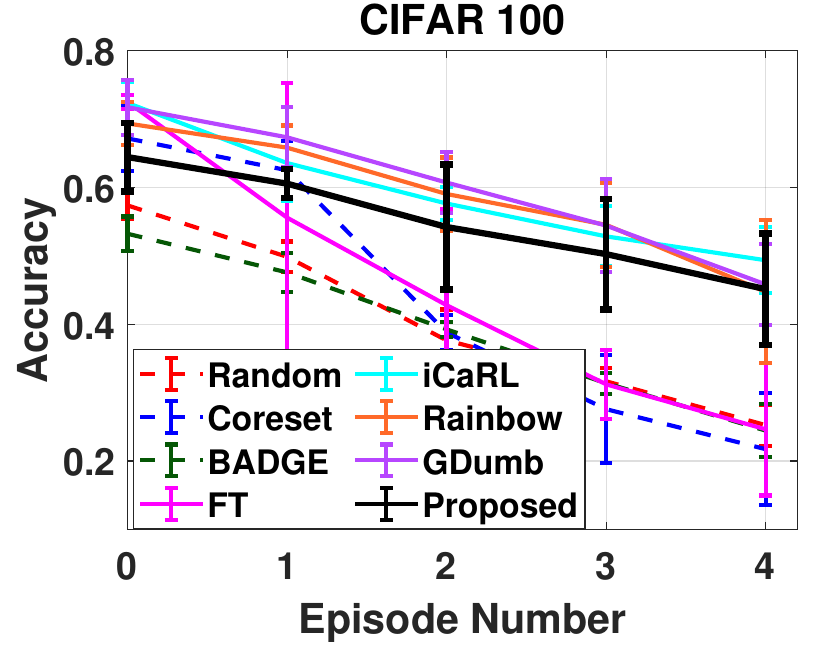}}
     \hspace{.02in}
		 \subfigure[Tiny ImageNet]{
          \label{fig_tinyimagenet}
          \includegraphics[trim = 1.3in 3.2in 1.7in 3.4in,clip,width=.31\textwidth]{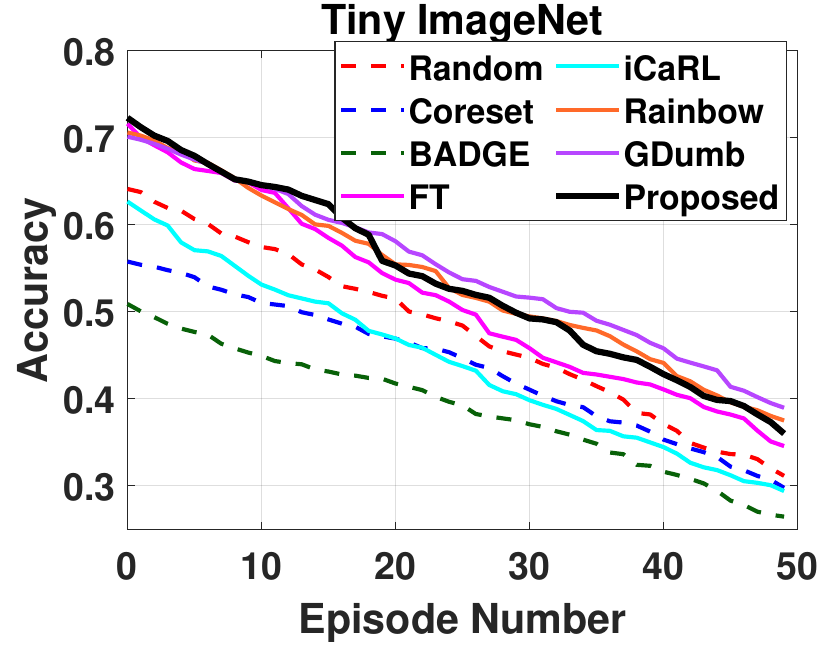}}    
                         
	\caption{Performance analysis of \textit{ACIL}. The AL baselines (\textit{Random, Coreset, BADGE}) are shown with dotted lines; the CIL baselines (\textit{Finetuning, iCaRL, Rainbow, GDumb}) and the proposed \textit{ACIL} method are shown with solid lines. The error bars have been omitted from the Tiny ImageNet results for better visualization. Best viewed in color.}
        \label{fig_main_results}     
\end{figure*}

The learning performance results are depicted in Figure \ref{fig_main_results} (accuracy) and Table \ref{tab_main_results_annotation} (annotation effort). In each graph in Figure \ref{fig_main_results}, the $x$-axis denotes the episode number, and the $y$-axis denotes the accuracy. Each plot represents the mean results of three runs to rule out the effects of randomness. Since the deep model is continually updated with new class information, the accuracy decays as more classes are incorporated with increasing episodes. 

\textbf{Accuracy: \textit{ACIL} vs. AL baselines:} \textit{ACIL} comprehensively outperforms all the AL baselines, and depicts much better accuracy. This is because, even though the AL baselines select an exemplar set in each episode, the samples are selected completely from the unlabeled set of the corresponding episode, as proposed in \cite{belouadah2020active}; thus, they fail to capture the knowledge from the former episodes and hence are not effective in mitigating catastrophic forgetting. 

\textbf{Accuracy: \textit{ACIL} vs. CIL baselines:} \textit{ACIL} depicts comparable performance as the CIL baselines, \textit{iCaRL}, \textit{GDumb} and \textit{Rainbow}. Our framework selects an exemplar set in each episode, based on an uncertainty and diversity based criterion, and is thus able to retain useful information about the former episodes, which enables it to mitigate catastrophic forgetting. Thus, the accuracy drops at more or less the same rate as the CIL baselines, with increasing number of episodes. \textit{Finetuning} depicts much worse performance, as it does not use an exemplar set to retain knowledge from previous episodes. 

We note that \textit{ACIL} sometimes depicts slightly better learning performance than the CIL baselines (where all the samples are labeled in all the episodes); for instance, it is better than \textit{iCaRL} for Tiny ImageNet. \textit{iCaRL} selects the exemplar set iteratively such that the mean feature vector of the exemplar is a good approximation of the mean feature vector of all the training samples in a given episode. This does not necessarily guarantee a good set of exemplars. For instance, it is possible to select a batch of uninformative / redundant samples whose mean is very close to the mean of all the samples. \textit{ACIL} on the other hand, selects samples by optimizing a criterion based on uncertainty and diversity, and thus selects a better set of exemplars to append to the next episode. Thus, even though all the samples are labeled in all the episodes in \textit{iCaRL}, by selecting a better set of exemplars to propagate from episode to episode, \textit{ACIL} is able to obtain a better learning performance. 

\textit{ACIL} also performs better than \textit{GDumb} in the later episodes of CIFAR 10 and COIL. \textit{GDumb} selects the exemplars to ensure that the class distributions are balanced; it does not consider the quality / informativeness of the exemplars. \textit{ACIL} propagates a more informative set of exemplars to the subsequent episode, which explains its marginally better performance. 

\begin{table*}[h]
	\centering
	\scriptsize
\begin{tabular}{|c|c|c|c|c|}
\hline
\textbf{} & \textbf{CIL Baselines} & \textbf{AL Baselines} & \textbf{Proposed}  \\
\hline
%\textbf{Method} & \textbf{Random} & \textbf{DPP} & \textbf{Submod} & \textbf{Proposed}  \\
%\hline
\textbf{MNIST} & $12,000 \pm 478.12$ & $3,359.6 \pm 483.43$ & $2,899.6 \pm 447.38$  \\
\hline
\textbf{CIFAR 10} & $10,000 \pm 0.00$ & $2,800 \pm 414.04$ & $2,417.2 \pm 344.90$  \\
\hline
\textbf{SVHN} & $14,651.40 \pm 5,542.22$ & $4,530 \pm 1,181.98$ & $3,763.60 \pm 1,455.34$ \\
\hline
\textbf{COIL} & $570 \pm 0.00$ & $204 \pm 30.51$ & $153.33 \pm 25.38$ \\
\hline
\textbf{CIFAR 100} & $5,000 \pm 0.00$ & $1,450 \pm 152.56$ & $1,158 \pm 131.53$ \\
\hline
\textbf{Tiny ImageNet} & $1,000 \pm 0.00$ & $299 \pm 9.97$ & $232.07 \pm 21.92$  \\
\hline
\end{tabular}
\caption{Average ($\pm$ std) number of samples that needed to be annotated per episode (including the episodic labeled set $X^{L}_n$) by the CIL baselines (\textit{Finetuning, iCaRL, Rainbow, GDumb}), AL baselines (\textit{Random, Coreset, BADGE}) and the proposed method (\textit{ACIL}).}
	\label{tab_main_results_annotation}
\end{table*}

\textbf{Annotation effort:} Table \ref{tab_main_results_annotation} reports the average number of samples that had to be annotated per episode (including the episodic labeled set $X^{L}_{n}$) for all the methods. The CIL baselines require all the samples to be annotated in each episode, which incurs substantial annotation effort. The AL baselines depict much less annotation effort, as only $k$ unlabeled samples need to be annotated in each episode. \textit{ACIL} depicts the least annotation effort, as it splits the budget $k$ to select samples from the exemplar set $E_{n-1}$ and the unlabeled set $X^{U}_{n}$ in episode $n$; since the exemplar set is already annotated, only the samples selected from $X^{U}_{n}$ need to be annotated in each episode for \textit{ACIL}. For the CIFAR 10 dataset, for instance, \textit{ACIL} results in about $4.13$ fold reduction in annotation effort on average in each episode, compared to the CIL baselines, while furnishing performance very similar to these baselines. Thus, \textit{ACIL} depicts comparable (and sometimes marginally better) performance than the CIL baselines at substantially reduced annotation effort; it also depicts much better performance than the AL baselines, at slightly reduced annotation effort. \textit{ACIL} is thus a promising solution for real-world incremental learning applications. 
\begin{table*}[h]
	\centering
	%\footnotesize
	\tiny
\begin{tabular}{|c|c|c|c|c|c|c|c|c|}
\hline
 \textbf{} & \textbf{Random} & \textbf{Coreset} & \textbf{BADGE} & \textbf{FT} & \textbf{iCaRL} & \textbf{Rainbow} & \textbf{GDumb} & \textbf{Proposed} \\  
\hline
 \textbf{MNIST} & $0.41 \pm 0.03$ & $0.39 \pm 0.01$ & $0.41 \pm 0.02$ & $0.20 \pm 0.01$ & $0.96 \pm 0.01$ & $\underline{0.97} \pm 0.01$ & $0.96 \pm 0.01$ & $\textbf{0.98} \pm 0.02$ \\   
\hline
 \textbf{CIFAR 10} & $0.37 \pm 0.03$ & $0.29 \pm 0.03$ & $0.27 \pm 0.17$ & $0.18 \pm 0.04$ & $0.42 \pm 0.06$ & $\textbf{0.5} \pm 0.02$ & $0.43 \pm 0.01$ & $\underline{0.47} \pm 0.02$ \\  
\hline
 \textbf{SVHN} & $0.49 \pm 0.09$ & $0.33 \pm 0.1$ & $0.22 \pm 0.02$ & $0.23 \pm 0.08$ & $\underline{0.70} \pm 0.02$ & $\textbf{0.75} \pm 0.03$ & $\textbf{0.75} \pm 0.01$ & $\textbf{0.75} \pm 0.04$ \\  
\hline
 \textbf{COIL} & $0.24 \pm 0.05$ & $0.17 \pm 0.02$ & $0.11 \pm 0.02$ & $0.02 \pm 0.02$ & $0.32 \pm 0.01$ & $\textbf{0.56} \pm 0.04$ & $\underline{0.5} \pm 0.03$ & $\textbf{0.56} \pm 0.02$ \\ 
\hline
 \textbf{CIFAR 100} & $0.25 \pm 0.03$ & $0.22 \pm 0.08$ & $0.24 \pm 0.04$ & $0.25 \pm 0.10$ & $\textbf{0.49} \pm 0.05$ & $0.45 \pm 0.10$ & $\underline{0.46} \pm 0.06$ & $0.45 \pm 0.08$ \\  
\hline
 \textbf{Tiny ImageNet} & $0.31 \pm 0.01$ & $0.30 \pm 0.03$ & $0.26 \pm 0.04$ & $0.35 \pm 0.02$ & $0.29 \pm 0.02$ & $\underline{0.38} \pm 0.05$ & $\textbf{0.39} \pm 0.04$ & $0.36 \pm 0.04$  \\ 
\hline
\end{tabular}
\caption{Average ($\pm$ std) accuracy after the last episode achieved by the CIL baselines (\textit{Finetuning, iCaRL, Rainbow, GDumb}), AL baselines (\textit{Random, Coreset, BADGE}) and the proposed method (\textit{ACIL}). Best results are marked in \textbf{bold} and second best results are \underline{underlined}.}
	\label{tab_last_accuracy}
\end{table*}

Table \ref{tab_last_accuracy} reports the accuracy obtained after the last episode for all the methods. We note that the proposed method achieves the highest accuracy in $3$ out of the $6$ datasets and the second highest accuracy in $1$ dataset. These results further corroborate the fact that \textit{ACIL} depicts competitive performance as the CIL baselines, at much reduced annotation effort. 

\subsection{Study of the exemplar set size (query budget)}

\begin{figure}[h]
	\centering
		\subfigure[Budget 500]{
          \label{fig_budget_500}
          \includegraphics[trim = 1.3in 3.2in 1.7in 3.4in,clip,width=.3\textwidth]{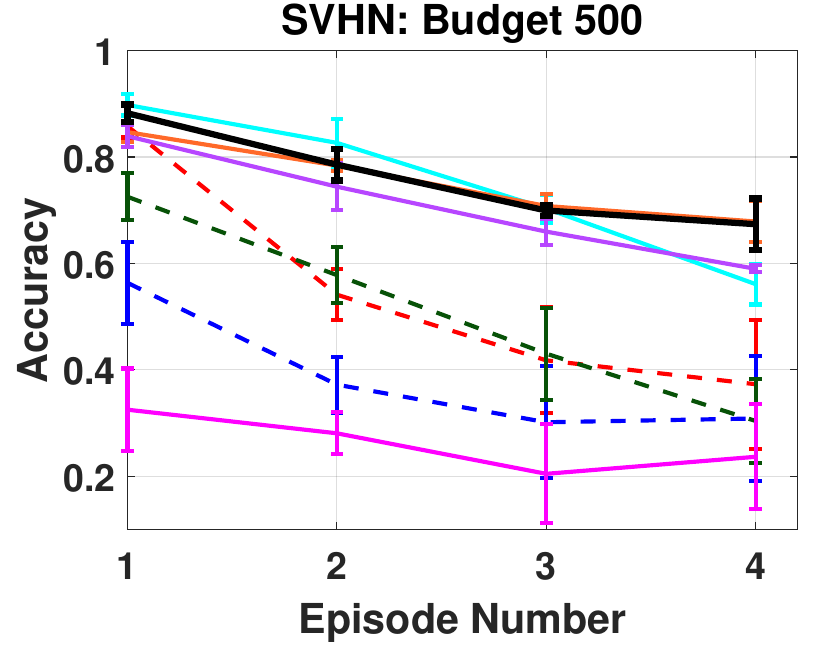}}  
     \hspace{.03in}
     \subfigure[Budget 1000]{
          \label{fig_budget_1000}
          \includegraphics[trim = 1.3in 3.2in 1.7in 3.4in,clip,width=.3\textwidth]{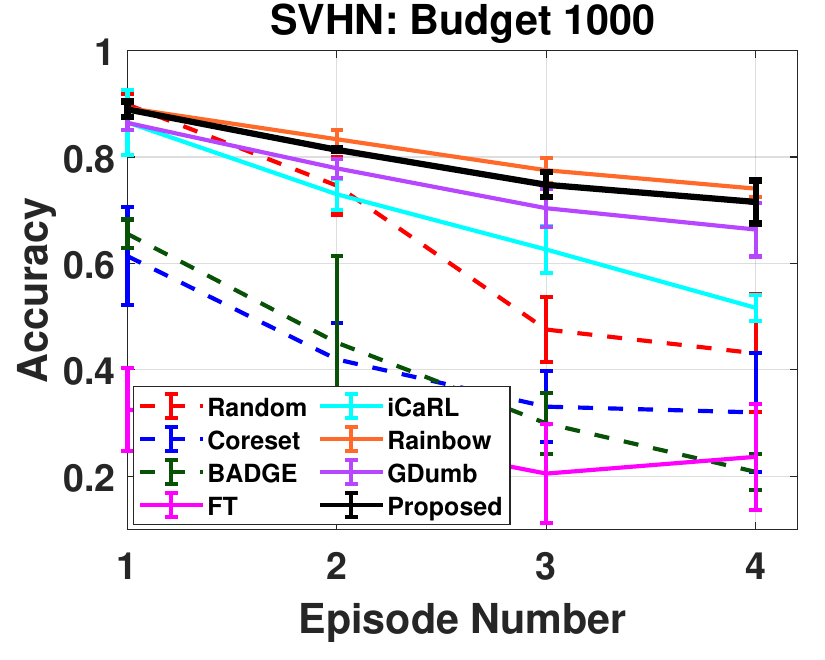}} 
     \hspace{.03in}
		 \subfigure[Budget 2500]{
          \label{fig_budget_2500}
          \includegraphics[trim = 1.3in 3.2in 1.7in 3.4in,clip,width=.3\textwidth]{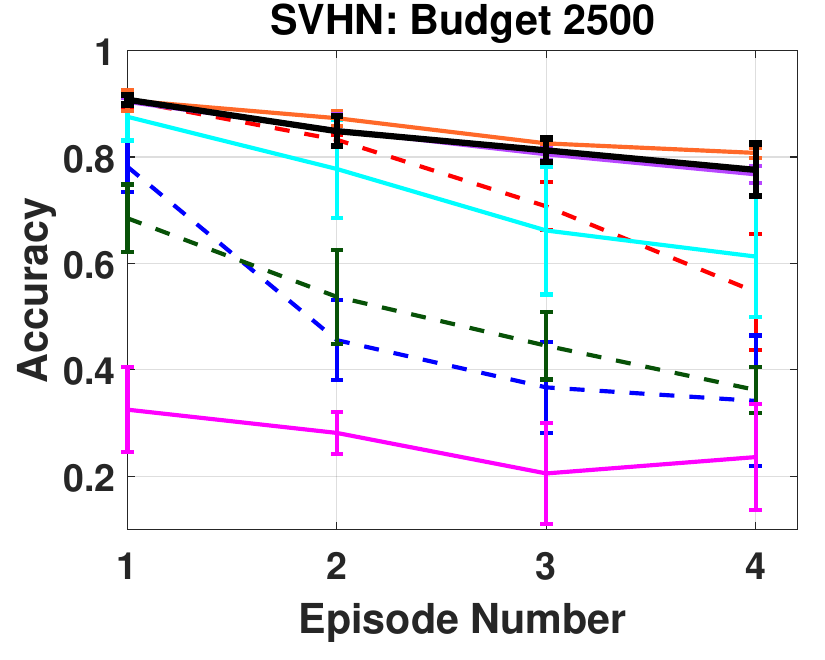}}   
             				 		
	\caption{Study of the exemplar set size (budget) on the SVHN dataset. The AL baselines (\textit{Random, Coreset, BADGE}) are shown with dotted lines; the CIL baselines (\textit{Finetuning, iCaRL, Rainbow, GDumb}) and the proposed \textit{ACIL} method are shown with solid lines. Best viewed in color.}
	\label{fig_budget_results}
\end{figure}

The goal of this experiment was to study the effect of the exemplar set size (budget) on the learning performance. The results on the SVHN dataset with budgets $500$, $1,000$ and $2,500$ are depicted in Figure \ref{fig_budget_results} (accuracy) and Table \ref{tab_budget_results} (annotation effort) (the default budget in Figure \ref{fig_svhn} was $2,000$). We note that the accuracy shows an increasing trend with an increase in budget, which is intuitive. A similar pattern is evident for all the budgets, where \textit{ACIL} depicts accuracy comparable to the CIL baselines (\textit{iCaRL}, \textit{GDumb} and \textit{Rainbow}) and much better than the AL baselines. 

From Table \ref{tab_budget_results}, it is also evident that \textit{ACIL} results in substantial savings in the human annotation effort compared to the CIL methods. The CIL baselines need all the samples to be annotated in all the episodes, so their annotation cost is not affected by the budget of the exemplar set. \textit{ACIL}, on the other hand, distributes the available budget between the unlabeled samples in a given episode and the exemplar set obtained from the previous episode. The results demonstrate that \textit{ACIL} depicts comparable performance to the CIL baselines at substantially reduced human annotation effort; it depicts much better performance than the AL baselines at slightly reduced annotation effort. These results depict the robustness of \textit{ACIL} to the budget of the exemplar set. This result is particularly important from a practical standpoint, since the budget of the exemplar set is dependent on the available memory / resources of a given application, and is different for different applications. 

\begin{table}[h]
	\centering
	\scriptsize
\begin{tabular}{|c|c|c|c|c|}
\hline
\textbf{} & \textbf{CIL Baselines} & \textbf{AL Baselines} & \textbf{Proposed}  \\
\hline
%\textbf{Method} & \textbf{Random} & \textbf{DPP} & \textbf{Submod} & \textbf{Proposed}  \\
%\hline
\textbf{SVHN: Budget 500} & $14,651.40 \pm 5,542.22$ & $3,330 \pm 1,068.78$ & $3,139.20 \pm 1,166.54$  \\
\hline
\textbf{SVHN: Budget 1,000} & $14,651.40 \pm 5,542.22$ & $3,730 \pm 1,068.41$ & $3,347.20 \pm 1,246.61$  \\
\hline
\textbf{SVHN: Budget 2,500} & $14,651.40 \pm 5,542.22$ & $4,930 \pm 1,285.86$ & $3,972.80 \pm 1,577.21$  \\
\hline
\end{tabular}
\caption{Average number of samples that needed to be annotated per episode (including the episodic labeled set $X^{L}_n$) by the CIL baselines (\textit{Finetuning, iCaRL, Rainbow, GDumb}), AL baselines (\textit{Random, Coreset, BADGE}) and the proposed method (\textit{ACIL}) for different budgets of the exemplar set on the SVHN dataset}
	\label{tab_budget_results}
\end{table}

\textit{We also conducted the following experiments and the results are reported in the Supplemental File: $(i)$ study of the backbone network architecture; $(ii)$ learning performance with varying number of episodes; $(iii)$ an analysis of the computation time of all the methods studied; $(iv)$ ablation studies to analyze the importance of the uncertainty and diversity components in our framework; and $(v)$ performance analysis of our framework using the retention metric.}

\section{Conclusion and Future Work}
\label{sec_conc}

In this paper, we proposed \textit{ACIL}, a novel active learning framework for class incremental learning settings. Contrary to existing CIL techniques, \textit{ACIL} does not require all the samples in each episode to be annotated; it selects an exemplar set in each episode, which is annotated and appended to the data in the next episode, so that the deep neural network trained in a given episode can retain knowledge of the classes encountered in past episodes. We formulated a criterion based on uncertainty and diversity to select the exemplar set in each episode; we also proposed a budget splitting strategy to ensure the exemplar set captures informative samples from all classes seen in former episodes. Our extensive empirical analyses on six vision datasets corroborated that \textit{ACIL} can not only mitigate catastrophic forgetting and deliver accuracy comparable to state-of-the-art CIL methods, but can also result in substantial savings of human annotation effort. As part of future work, we plan to extend our framework to address IL in the regression setup, which has attracted research attention recently \cite{Continual_Reg}. 

\section{Acknowledgment}
\label{sec_ack}

This research was supported in part by the National Science Foundation under Grant Number: IIS-2143424 (NSF CAREER Award).

\bibliography{BMVC_2024_ACIL_bib}
\end{document}